\documentclass[runningheads]{llncs}
\usepackage[T1]{fontenc}
\usepackage{graphicx}
\usepackage{wrapfig}
\usepackage{array}
\usepackage{tabularx}
\usepackage{booktabs}
\usepackage{booktabs, multirow} 
\usepackage{subcaption} 

\begin{document}
\title{Context-Aware Pesticide Recommendation via Few-Shot Pest Recognition for Precision Agriculture}
%
%
\author{Anirudha Ghosh\inst{1} \and
        Ritam Sarkar\inst{2} \and
        Debaditya Barman\inst{1,}\thanks{Corresponding author. Email: debaditya.barman@visva-bharati.ac.in}}
\institute{Department of Computer and System Sciences, Visva-Bharati, Santiniketan, India \and
           Faculty of Technology, Uttar Banga Krishi Viswavidyalaya, Cooch Behar, India}

\authorrunning{A. Ghosh et al.}
%
\maketitle              
\begin{figure}[]
	\begin{center}
   		\includegraphics[width=0.92\textwidth]{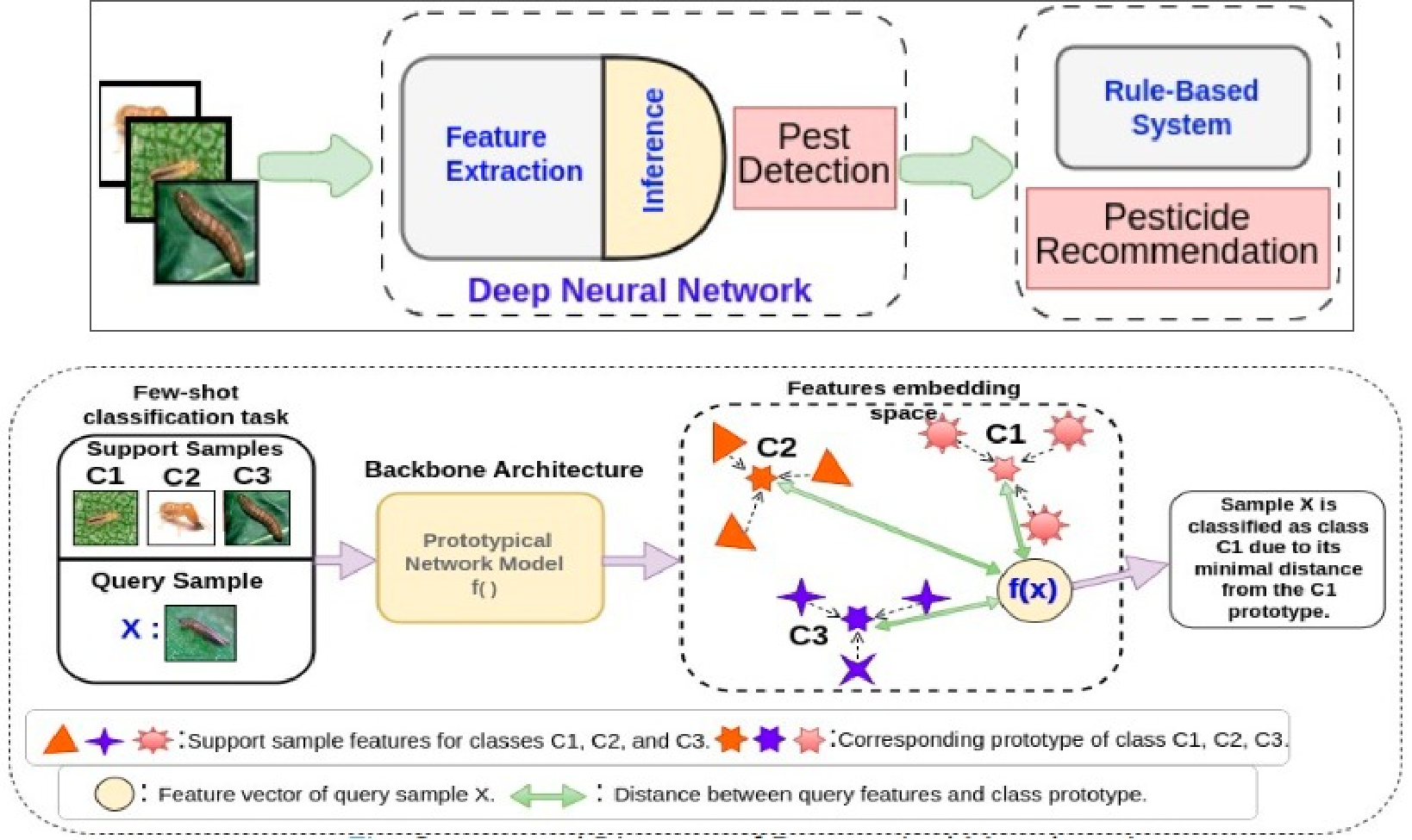}
	\end{center}
	\caption{Overview of our proposed framework integrating a lightweight prototypical network for few-shot pest detection with the rule-based Decision Support System (integrates pest detection results with environmental factors) to recommend context-aware pesticides.}
	\label{fig:MainCNN}
    \vspace{-1cm} 
\end{figure}
\begin{abstract}
Effective pest management is crucial for enhancing agricultural productivity, especially for crops such as sugarcane and wheat that are highly vulnerable to pest infestations. Traditional pest management methods depend heavily on manual field inspections and the use of chemical pesticides. These approaches are often costly, time-consuming, labor-intensive, and can have a negative impact on the environment. To overcome these challenges, this study presents a lightweight framework for pest detection and pesticide recommendation, designed for low-resource devices such as smartphones and drones, making it suitable for use by small and marginal farmers.

The proposed framework includes two main components. The first is a Pest Detection Module that uses a compact, lightweight convolutional neural network (CNN) combined with prototypical meta-learning to accurately identify pests even when only a few training samples are available. The second is a Pesticide Recommendation Module that incorporates environmental factors like crop type and growth stage to suggest safe and eco-friendly pesticide recommendations.
To train and evaluate our framework, a comprehensive pest image dataset was developed by combining multiple publicly available datasets. The final dataset contains samples with different viewing angles, pest sizes, and background conditions to ensure strong generalization.

Experimental results show that the proposed lightweight CNN achieves high accuracy, comparable to state-of-the-art models, while significantly reducing computational complexity. The Decision Support System additionally improves pest management by reducing dependence on traditional chemical pesticides and encouraging sustainable practices, demonstrating its potential for real-time applications in precision agriculture.

\keywords{Pest Detection \and Pesticide Recommendation \and Deep Learning \and Pest Management \and Prototypical Meta-Learning \and Few-Shot Classification \and Edge Computing \and Sustainable Agriculture \and Precision Farming.}
\end{abstract}

\section{Introduction}
Pests cause substantial economic losses by damaging crops, with sugarcane \cite{usman2020sugarcane} and wheat \cite{sehgal2021management} particularly vulnerable due to their across-the-board cultivation and economic significance. Pest infestations like cutting weevils, termites, and leafcutter ants in sugarcane, or wheat rust and cutworms in wheat, can drastically decrease outcomes. These losses not only affect farmers but also disrupt supply chains and threaten food availability for millions of people worldwide. So, effective pest management is vital for maintaining agricultural productivity with protecting global food security \cite{pedigo2021entomology}. 

Traditional pest management practices, though adequate to some extent, face multiple challenges. Manual assessment for pest detection requires significant time investment and skilled labor, making it impractical for large-scale agriculture functions. Moreover, the overreliance on chemical pesticides, although delivering quick results, has led to long-term health and environmental damage \cite{kandalkar2014classification}. These include water pollution, soil degradation \cite{larios2008automated}, pest pesticide antagonism, and biodiversity loss \cite{messelink2021biodiversity}. Therefore, there is a need for efficient, sustainable, and scalable pest management solutions to manage these growing challenges.

To overcome the above challenges, this study focuses on developing a lightweight framework for pest detection and pesticide recommendations explicitly for sugarcane and wheat crops. The framework is designed to operate effectively with limited data (few training samples) by leveraging prototypical meta-learning \cite{wang2021meta} combined with a few-shot learning \cite{parnami2022learning} strategy. This strategy allows quick model adaptation to different pest types while maintaining good accuracy. Furthermore, combining a decision support system enables context-aware pesticide recommendations, promotes eco-friendly practices, and minimizes the chemical usage.

The key contributions of this work comprise the creation of a comprehensive pest dataset for sugarcane and wheat, a robust lightweight deep neural network for pest detection, and a rule-based recommendation system for pesticide recommendation. The framework highlights practicality, being compatible with low-resourced edge devices like drones and smartphones, allowing farmers to aid from real-time pest management tools. This study highlights the prospect of combining machine learning techniques with sustainable agricultural techniques to tackle demanding pest management challenges.

\section{Related Work}
Traditional pest management strategies depend on manual inspections and chemical pesticide applications. While useful in some situations, these methods are time-consuming, labor-intensive, and unsuited for large-scale agriculture operations. The overreliance on chemical pesticides has resulted in environmental degradation, pesticide resistance, and health risks \cite{kandalkar2014classification,larios2008automated}. Recent advancements in deep learning have revolutionized pest classification to address these challenges, providing accurate and scalable solutions for precision agriculture.

Convolutional Neural Networks (CNNs) have appeared as the dominant approach for this pest detection and classification problem. The work in \cite{liu2020deep}, developed a hybrid global and local feature-based pest monitoring system, acquiring high accuracy in multi-class pest detection. Similarly, the study at \cite{nanni2020insect}, integrated CNN models such as AlexNet and DenseNet201 with saliency-based methods to classify pest images, documenting an accuracy of 92.43\% on a smaller dataset and 61.93\% on the IP102 dataset. The study done by \cite{li2019effective} used Region Proposal Networks (RPNs) and data augmentation strategies for pest localization, gaining a mean Average Precision (mAP) of 83.23\%. In \cite{roy2022fast}, authors presented a fine-grained detection model based on YOLOv4, acquiring a mAP of 96.29\%, showcasing the robustness of CNN-based methods in agricultural pest detection.

Recently, lightweight CNNs have been presented to address the computational constraints of deploying models on resource-low devices like drones and smartphones. For instance, \cite{setiawan2022large} optimized MobileNetV2 using dynamic learning rates with proper augmentation, achieving 71.32\% accuracy on the IP102 dataset. The study done by \cite{xia2018insect}, used DenseNet169 with transfer learning, reaching 88.83\% accuracy for tomato pest classification. These lightweight models significantly reduce computational overhead, making them suitable for real-time applications. But their dependence on large datasets restricts their ability to generalize to rarely collected pests with unstructured environments.

Few-shot learning (FSL) has recently emerged as a solution to overcome training data scarcity. FSL allows models to generalize to unknown classes with limited labeled data. In the study by \cite{dhillon2019baseline}, the authors showed improved accuracy in one-shot scenarios by using fine-tuning techniques. In the survey \cite{wang2020frustratingly}, the authors employed lightweight architectures such as MobileNet for few-shot pest detection, achieving considerable performance advancements. The work of \cite{nieuwenhuizen2018detection} explored metric learning strategies for few-shot pest classification, acquiring results in detecting cotton pests under several real-world scenarios. 

Despite advancements in pest detection, integrating pest detection systems with actionable pesticide suggestions still needs to be investigated. The work by \cite{li2022classification} showed multi-class pest detection utilizing ResNet152 but lacked context-aware pesticide suggestions. Likewise, \cite{nieuwenhuizen2018detection} earned high detection accuracy utilizing Faster R-CNN but still required integrating a decision support system. This study presents a lightweight CNN integrated with a Decision Support System (DSS) to address these gaps. 
\section{Methodology}
\subsection{Framework Overview}
Our proposed framework consists of two primary modules developed to enable efficient and real-time pest management for sugarcane and wheat crops:
\begin{itemize}
    \item \textbf{Pest Detection Module: }This module employs a lightweight Convolutional Neural Network (CNN) for real-time pest detection. The model analyzes images captured from fields and accurately classifies pests into predefined categories using minimal data by leveraging prototypical meta-learning with a few-shot strategy. This lightweight design ensures compatibility with resource-constrained devices like smartphones and drones, enabling on-the-go pest identification.
    \item \textbf{Pesticide Recommendation Module: }A rule-based Decision Support System (DSS) complements the pest detection module by combining pest classification results with environmental aspects such as crop type, growth stage, and pest severity. The DSS uses a knowledge mapping table to recommend suitable pesticides while emphasizing eco-friendly and decreased chemical usage procedures. This integration ensures context-aware, sustainable pest management solutions.    
\end{itemize}
\subsection{Dataset Creation}
The dataset for sugarcane and wheat pests was created using a structured process to ensure balance and quality, with 30 samples per pest class from copyright-free repositories. Images were collected for pests such as Cutting Weevil, Leafcutter Ants, Red Palm Weevil, Sugarcane Woolly Aphids, and Termites for sugarcane, and Cutworms, Wheat Stem Sawfly, Termites, Wheat Thrips, and Wheat Leaf Rust for wheat. Pre-processing included resizing images to $100\times100$ pixels and ensuring quality and uniformity across all samples. A research team processed the dataset manually to eliminate irrelevant samples and maintain quality. Organized hierarchically by crop and pest categories, the final dataset includes 300 images, with 30 samples per class, ensuring sufficient diversity and readiness for testing through lightweight CNNs.
\subsection{Model Architecture}
The pest detection module employs a lightweight CNN architecture optimized for edge devices, utilizing prototypical meta-learning to enable effective pest identification with minimal data in few-shot settings. This design ensures compatibility with resource-constrained environments.

\begin{figure}[]
	\begin{center}
   		\includegraphics[width=0.8\textwidth]{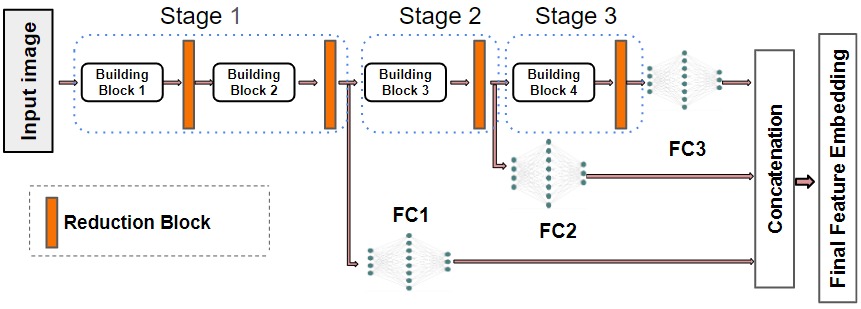}
	\end{center}
	\caption{Architecture of propose Lightweight Prototype Network Backbone.}
	\label{fig:MainCNN}
\end{figure}
The Lightweight Prototype Network Backbone (LPN-Backbone) architecture, illustrated in Figure \ref{fig:MainCNN}, is developed with significantly fewer learnable parameters compared to ResNet18 and DenseNet169. While ResNet18 and DenseNet169 have approximately 12 million and 14 million parameters, the  LPN consists of roughly 10 million. The LPN is structured into three stages for hierarchical feature extraction. It comprises four Feature Extraction Building Blocks (FEBBs) and four Reduction Blocks (RBs). The first stage includes two FEBBs, each followed by an RB, while the second and third stages feature one FEBB and one RB each. At the end of each stage, a Fully Connected module (FC) with three densely connected layers combines extracted features into the final feature embedding.
\begin{wrapfigure}[8]{r}{0.48\textwidth}
  \centering
  \includegraphics[width=0.48\textwidth]{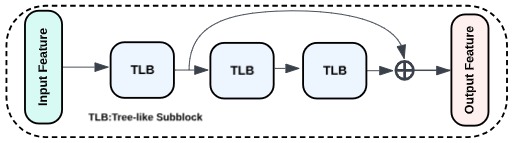}
  \caption{Architecture of FEBB}
  \label{fig:BB}
\end{wrapfigure}
Each FEBB, shown in Figure \ref{fig:BB}, is made up of three Targeted Learning Blocks (TLBs) with a residual connection linking the input of the second Targeted Learning Block to the output of the third Targeted Learning Block, addressing vanishing gradient issues and improving stability during training. Following each FEBB, the RB reduces feature dimensions using a (3×3) convolutional layer combined with batch normalization (BN) and ReLU activation.

The core of each FEBB is the TLB, which extracts diverse features using three independent branches, each using unique convolutional operations:
\begin{itemize}
    \item \textbf{Branch A:} Uses a ($3\times3$) depthwise separable convolution to extract spatial features, followed by a (1$\times$1) convolution to combine information across channels. Batch normalization (BN) and ReLU activation are applied to outputs consistent.
    \item \textbf{Branch B:} Inspired by ResNet, this branch starts with a (1$\times$1) convolution to reduce input size, followed by a (3$\times$3) grouped convolution to focus on essential features. A residual connection adds the input back to the output, ensuring feature preservation. BN and ReLU activation are applied to generate the final features with consistent shapes.
    \item \textbf{Branch C:} Combines a (1$\times$1) pointwise convolution to simplify channel dimensions with a (3$\times$3) dilated convolution to cover a wider input area. This design allows capturing more context without extra computations. BN and ReLU activation are applied to maintain stable outputs.
\end{itemize}

\subsection{Decision Support System (DSS)}

The DSS takes pest detection results along with environmental and growth-specific factors as input to recommend pesticide solutions for sugarcane and wheat. It relies on a rule-based knowledge to map pest types to optimal pesticide choices, as shown in Tables \ref{tab:suger} and \ref{tab:wheat}. As detailed in those tables, the DSS recommendations specify the kind of suitable pesticide for each pest, its corresponding crop, and environmental and growth stage conditions. However, dosage and timing are not explicitly listed in the tables and must be determined based on specific infestation severity and local guidelines.
\begin{table}[ht]
\centering
\caption{Knowledge Base for Pests Affecting Sugarcane}
\begin{tabularx}{\linewidth}{|X|X|X|X|}
\hline
\textbf{Pest} & \textbf{Growth Stage} & \textbf{Environmental Condition} & \textbf{Pesticide Recommended} \\ \hline
Cutting Weevil (\textit{Apion spp.}) & Early growth & High humidity & Mild insecticide (e.g., Chlorpyrifos) \\ \hline
Leafcutter Ants (\textit{Atta spp.}) & Vegetative & Warm and dry & Bait-based pesticides (e.g., Hydramethylnon) \\ \hline
Red Palm Weevil (\textit{Rhynchophorus ferrugineus}) & Any stage & High humidity & Pheromone-based lures and traps \\ \hline
Sugarcane Woolly Aphid (\textit{Siphunculus spp.}) & Vegetative & Moderate temperature, high humidity & Natural predators (lady beetles) or Imidacloprid \\ \hline
Termites (various species) & Any stage & Dry soil conditions & Soil treatment (e.g., Fipronil) \\ \hline
\end{tabularx}
\label{tab:suger}
\end{table}

\begin{table}[ht]
\centering
\caption{Knowledge Base for Pests Affecting Wheat}
\begin{tabularx}{\linewidth}{|X|X|X|X|}
\hline
\textbf{Pest} & \textbf{Growth Stage} & \textbf{Environmental Condition} & \textbf{Pesticide Recommended} \\ \hline
Cutworms (\textit{Agrotis spp.}) & Early growth & Cool, moist & Biological pesticide (e.g., Bacillus thuringiensis) \\ \hline
Wheat Stem Sawfly (\textit{Cephus cinctus}) & Vegetative & Dry & Pheromone traps or mild insecticides \\ \hline
Wheat Thrips (\textit{Thrips tabaci}) & Vegetative & Warm and dry & Sticky traps or Lambda-cyhalothrin \\ \hline
Wheat Leaf Rust (\textit{Puccinia triticina}) & Vegetative & High humidity & Fungicide (e.g., Propiconazole) \\ \hline
Termites (various species) & Any stage & Dry soil conditions & Soil-applied insecticides (e.g., Chlorpyrifos) \\ \hline
\end{tabularx}
\label{tab:wheat}
\end{table}

\section{Experimental Setup}
\subsection{Training}
\subsubsection{Training Dataset}
Our framework utilizes prototypical meta-learning for few-shot classification, which demands a large collection of training with mini-tasks to simulate diverse scenarios effectively. To meet this requirement, we combined three publicly available datasets from Kaggle to form a comprehensive training dataset. The first dataset, the Dangerous Farm Insects Dataset \cite{dalal_dangerous_insects}, includes images of 15 harmful insects commonly found in agricultural environments. The second dataset, the Agricultural Pests Image Dataset \cite{lanz_agricultural_pests}, includes 12 pest classes: ants, bees, beetles, caterpillars, and weevils. Lastly, the Crop Pest Dataset \cite{ghosh_crop_pest} contains images of four pest types: rice stem borer, green leafhopper, rice bug, and rice leaf roller. This diverse collection of images enables the creation of varied training tasks, providing the foundation for model learning and generalization with a few-shot learning strategy.
\subsubsection{Preprocessing and Augmentation}
The preprocessing and augmentation pipeline involved resizing images to 100×100 pixels, applying random color jitter (brightness, contrast, saturation, and hue), and performing affine transformations with random zoom up to 15\%. Then, we apply random horizontal and vertical flips (with 50\% probability each), and finally, we perform a grayscale conversion with three channels and normalization using ImageNet statistics.
\subsubsection{Training Procedure}
The training process involves the meta-training of two base networks, ResNet18 and DenseNet169, along the proposed LPN, using the previously described training dataset. The models were trained over 200 meta-tasks (episodes), with each episode comprising separate mini-tasks. Each mini-task was trained for 40 epochs to complete the meta-training phase.
During training, the k-way, n-shot protocol was used, where k (number of classes) and n (number of support samples per class) were randomly chosen between 5 to 10 and 1 to 15, respectively.
The Adam optimizer minimizes the training loss, with a learning rate (LR) of 0.002 for all the networks. A dropout rate of 30\% was applied to regularize all models and prevent overfitting. This training strategy enhanced the models ability to generalize effectively to new, unseen tasks.
\begin{figure}[]
	\begin{center}
   		\includegraphics[width=0.99\textwidth]{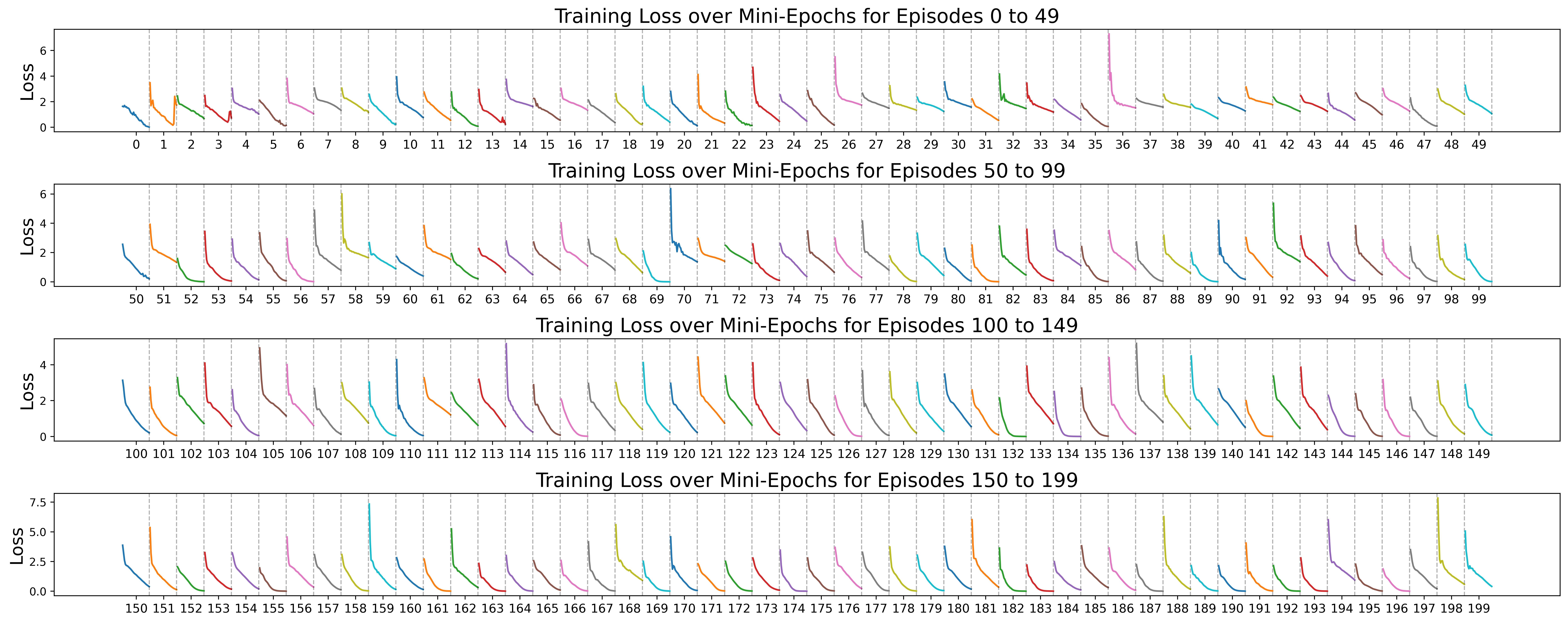}
	\end{center}
	\caption{Training loss of ResNet18.}
	\label{fig:resloss}
\end{figure}
\begin{figure}[]
	\begin{center}
   		\includegraphics[width=0.99\textwidth]{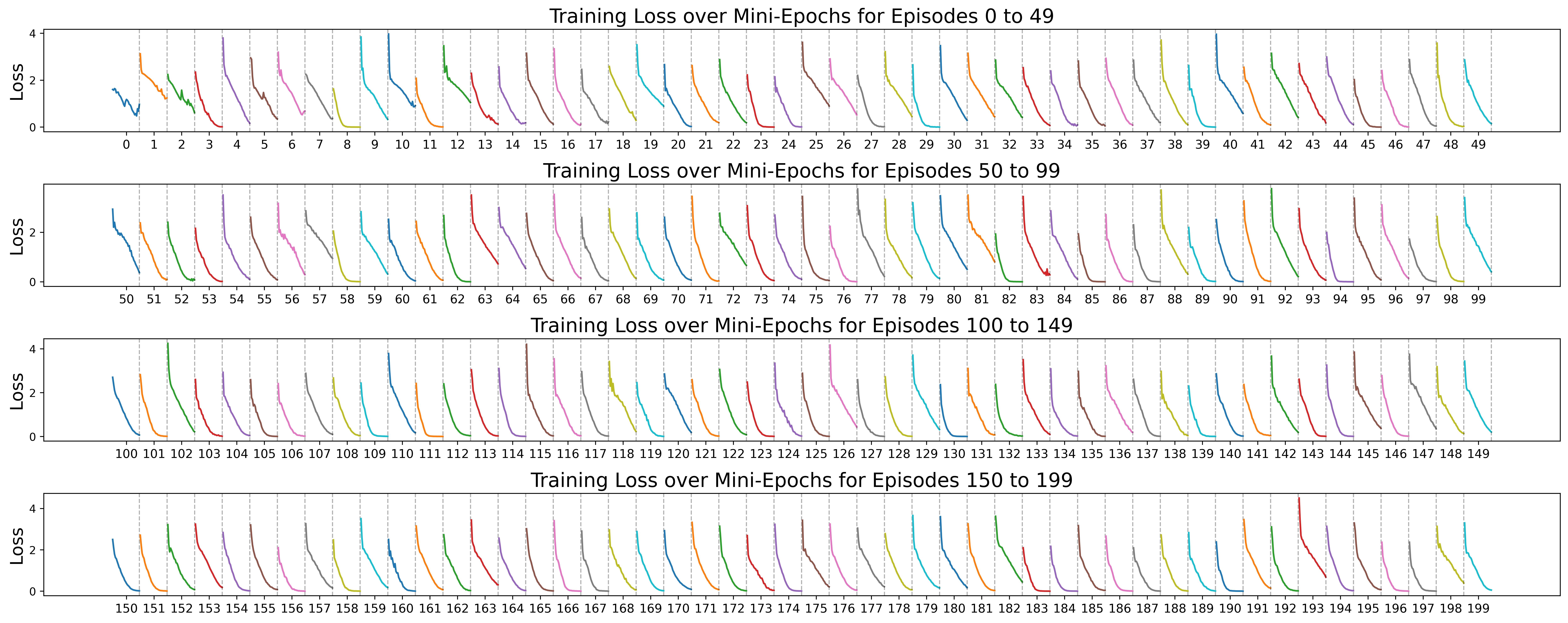}
	\end{center}
	\caption{Training loss of DensNet169.}
	\label{fig:lossdens}
\end{figure}
\begin{figure}[]
	\begin{center}
   		\includegraphics[width=0.99\textwidth]{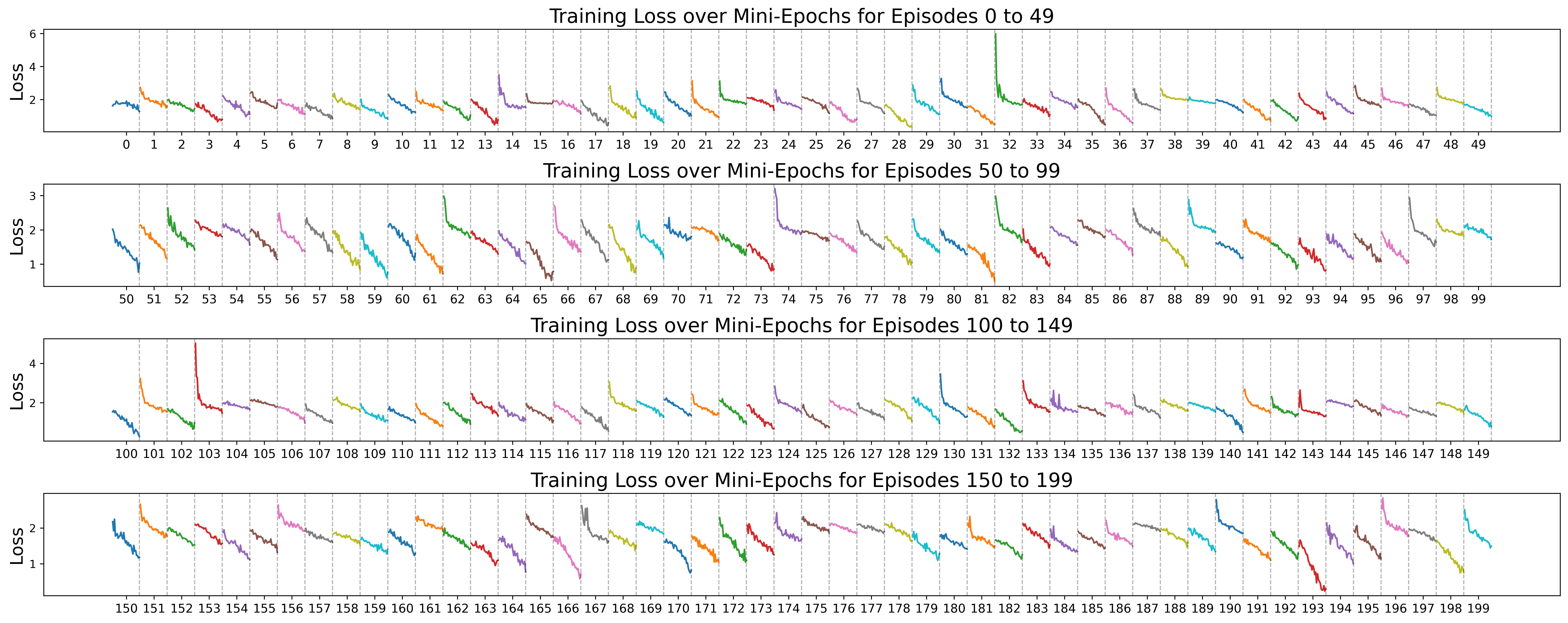}
	\end{center}
	\caption{Training loss of our proposed LPN.}
	\label{fig:lossown}
\end{figure}

Figure \ref{fig:resloss} illustrates the training loss for ResNet18, while Figure \ref{fig:lossdens} presents the training loss for DenseNet169. Similarly, Figure \ref{fig:lossown} depicts the training loss for the proposed LPN model, highlighting its convergence behavior during the meta-training phase.
\subsubsection{Testing Procedure}
The testing procedure utilize a few-shot strategy for both the sugarcane and wheat testing datasets, each comprising five classes. Testing was conducted under 5-way 1-shot, 3-shot, and 5-shot scenarios. For the 1-shot setting, one support image per class was randomly selected. The 3-shot scenario expanded this by including the 1-shot support image along with two additional random images per class. Similarly, the 5-shot setting built upon the 3-shot scenario by adding two more random support images per class.
\begin{figure}[!ht]
    \centering
    \begin{subfigure}[t]{0.49\linewidth}
        \centering
        \includegraphics[width=\linewidth]{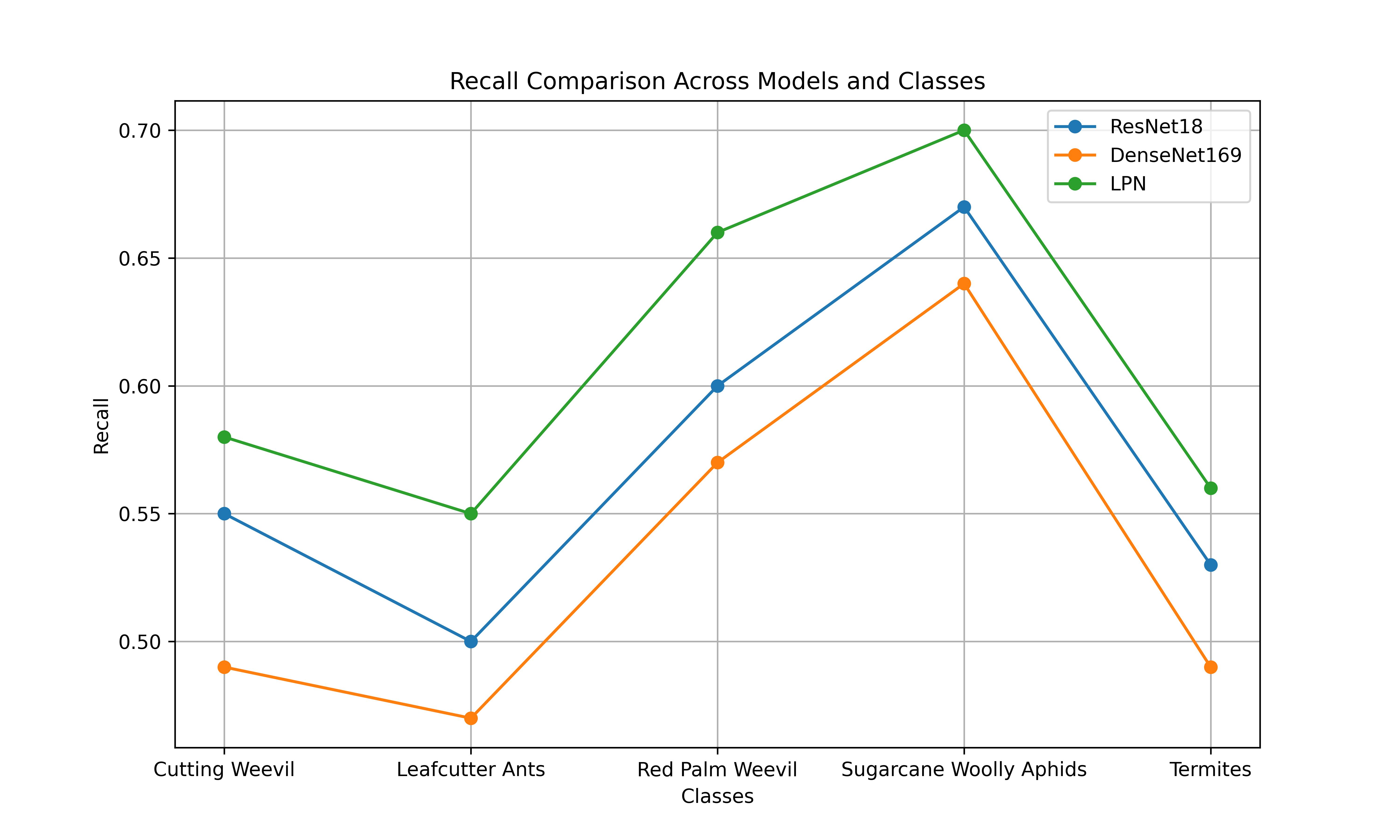}
        \caption{Recall Plot for Pest Detection Across 1-Shot, 3-Shot, and 5-Shot Configurations in Sugarcane}
        \label{fig:rps}
    \end{subfigure}
    \hfill
    \begin{subfigure}[t]{0.49\linewidth}
        \centering
        \includegraphics[width=\linewidth]{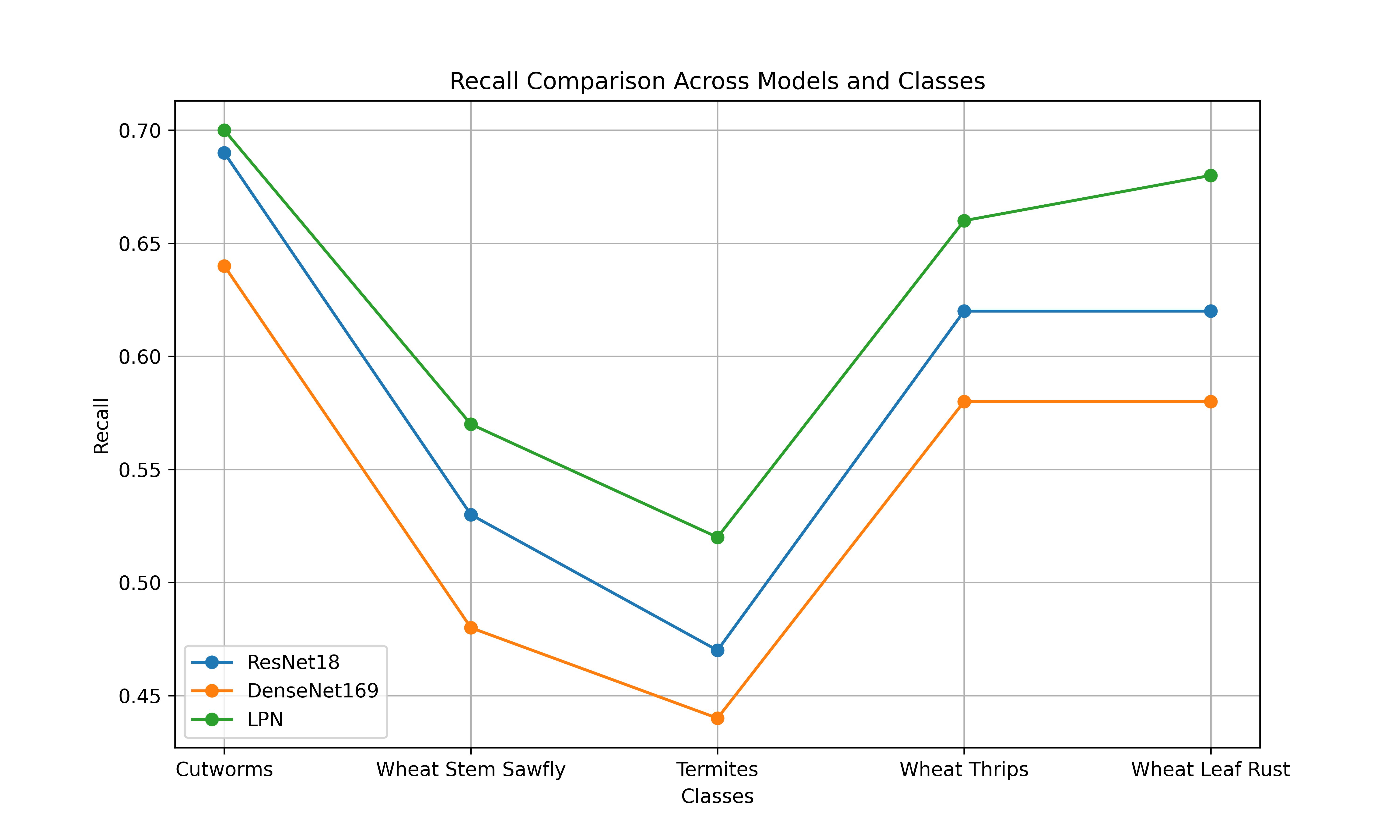}
        \caption{Recall Plot for Pest Detection Across 1-Shot, 3-Shot, and 5-Shot Configurations in Wheat}
        \label{fig:rpw}
    \end{subfigure}
    \caption{Comparison of Recall and Precision for Pest Detection Across Different Shot Configurations in Sugarcane and Wheat.}
    \label{fig:comparison_plots}
\end{figure}
\section{Results and Discussion}
\subsection{Performance Metrics}
The performance of all models for pest detection was evaluated using 1-shot, 3-shot, and 5-shot arrangements for wheat and sugarcane pests. The results, detailed in Tables \ref{tab:perfoaccSuger } and \ref{tab:acctabweath}, highlight the preeminent performance of our proposed LPN model compared to the baseline models (ResNet18 and DenseNet169). Each data point in those tables is generated based on the results of ten independent testing assignments. For wheat pests (Table \ref{tab:perfoaccSuger }), the LPN consistently exceeded the baselines in all configurations. In the 1-shot setting, the LPN achieved 72.00\% accuracy for Cutworms and 69.71\% for Wheat Leaf Rust, notably exceeding ResNet18 and DenseNet169. In the 5-shot configuration, the LPN further improved, showcasing its capability to leverage extra support images effectively.

Similarly, the LPN model displayed superior performance in sugarcane pest detection (Table \ref{tab:acctabweath}). In the 3-shot configuration, it achieved 69.44\% accuracy for Cutting Weevil and 77.62\% for Sugarcane Woolly Aphids, surpassing both ResNet18 and DenseNet169. In the 5-shot setting, the LPN model maintained its lead with 74.81\% accuracy for Red Palm Weevil and 68.43\% for Termites. 

Additionally, the pest detection recall plot (Figure \ref{fig:rps}) for sugarcane pest and (Figure \ref{fig:rpw}) for wheat pest further validate our models effectiveness, demonstrating consistent improvement in recall across various shot scenarios. These results highlight the robustness and adaptability of the LPN model, particularly in few-shot learning scenarios.
\begin{table}[!h]\centering
\caption{Performance Metrics (\% Accuracy) for Sugarcane Pest Detection}\label{tab:perfoaccSuger }
\tiny

\begin{tabularx}{\linewidth}{|X|X|X|X|X|X|X|X|}\hline
\textbf{Model} &\textbf{Shot Configuration} &\textbf{Cutting Weevil} &\textbf{Leafcutter Ants} &\textbf{Red Palm Weevil} &\textbf{Sugarcane Woolly Aphids} &\textbf{Termites} \\\cline{1-7}
\multirow{3}{*}{\textbf{ResNet18}} &1-Shot &55.11 ± 0.05 &50.21 ± 0.03 &61.23 ± 0.04 &67.48 ± 0.03 &53.21 ± 0.07 \\ 
&3-Shot &60.23 ± 0.06 &55.37 ± 0.04 &66.01 ± 0.05 &71.15 ± 0.04 &58.47 ± 0.06 \\ 
&5-Shot &64.78 ± 0.07 &59.84 ± 0.05 &67.54 ± 0.06 &73.89 ± 0.05 &61.95 ± 0.08 \\\cline{1-7}
\multirow{3}{*}{\textbf{DenseNet169}} &1-Shot &49.92 ± 0.06 &47.04 ± 0.04 &58.17 ± 0.05 &64.93 ± 0.04 &49.92 ± 0.08 \\ 
&3-Shot &55.12 ± 0.07 &52.09 ± 0.05 &62.43 ± 0.06 &68.31 ± 0.05 &54.35 ± 0.09 \\ 
&5-Shot &60.87 ± 0.08 &56.47 ± 0.06 &66.09 ± 0.07 &71.56 ± 0.06 &58.74 ± 0.10 \\\cline{1-7}
\multirow{3}{*}{\textbf{LPN}} &1-Shot &58.51 ± 0.05 &55.68 ± 0.03 &66.11 ± 0.04 &71.21 ± 0.03 &57.38 ± 0.07 \\ 
&3-Shot &64.87 ± 0.06 &61.22 ± 0.04 &72.35 ± 0.05 &74.97 ± 0.04 &63.19 ± 0.06 \\ 
&5-Shot &\textbf{69.44 ± 0.07} &\textbf{66.71 ± 0.05} &\textbf{74.81 ± 0.06} &\textbf{77.62 ± 0.05} &\textbf{68.43 ± 0.08} \\
\hline
\end{tabularx}

\end{table}

\begin{table}[!h]\centering
\caption{ Performance Metrics (\% Accuracy) for Wheat Pest Detection}\label{tab:acctabweath}
\tiny
\begin{tabularx}{\linewidth}{|X|X|X|X|X|X|X|X|}\hline
\textbf{Model} &\textbf{Shot Configuration} &\textbf{Cutworms} &\textbf{Wheat Stem Sawfly} &\textbf{Termites} &\textbf{Wheat Thrips} &\textbf{Wheat Leaf Rust} \\\cline{1-7}
\multirow{3}{*}{\textbf{ResNet18}} &1-Shot &69.92 ± 0.05 &54.75 ± 0.06 &48.69 ± 0.04 &64.16 ± 0.07 &63.14 ± 0.05 \\ 
&3-Shot &75.03 ± 0.06 &58.89 ± 0.07 &53.42 ± 0.05 &68.72 ± 0.08 &67.39 ± 0.06 \\ 
&5-Shot &78.56 ± 0.07 &63.21 ± 0.08 &57.98 ± 0.06 &72.93 ± 0.09 &71.64 ± 0.07 \\\cline{1-7}
\multirow{3}{*}{\textbf{DenseNet169}} &1-Shot &65.92 ± 0.05 &49.76 ± 0.06 &45.73 ± 0.04 &60.56 ± 0.07 &59.51 ± 0.05 \\ 
&3-Shot &70.14 ± 0.06 &53.88 ± 0.07 &50.21 ± 0.05 &64.98 ± 0.08 &63.18 ± 0.06 \\ 
&5-Shot &73.56 ± 0.07 &58.14 ± 0.08 &54.13 ± 0.06 &69.82 ± 0.09 &67.92 ± 0.07 \\\cline{1-7}
\multirow{3}{*}{\textbf{LPN}} &1-Shot &72.00 ± 0.05 &59.01 ± 0.06 &53.15 ± 0.04 &68.15 ± 0.07 &69.71 ± 0.05 \\ 
&5-Shot &76.58 ± 0.06 &63.87 ± 0.07 &57.61 ± 0.05 &72.48 ± 0.08 &73.23 ± 0.06 \\ 
&5-Shot &\textbf{80.12 ± 0.07} &\textbf{68.52 ± 0.08} &\textbf{61.19 ± 0.06} &\textbf{76.98 ± 0.09} &\textbf{75.96 ± 0.07} \\\hline

\end{tabularx}
\end{table}
\subsection{Analysis and Discussion}
The experimental results highlight the superior performance of the lightweight LPN model, which consistently outperforms ResNet18 and DenseNet169, especially in the 3-shot and 5-shot settings, showing its ability to utilize additional support samples to improve classification accuracy. A paired t-test confirms that these performance gains are statistically significant (p < 0.05) across both sugarcane and wheat pest datasets, establishing the reliability of the proposed approach. Its low computational needs make it ideal for real-time deployment on resource-constrained devices. Further, the Integration of the Decision Support System (DSS) improves the framework by offering actionable pesticide recommendations, prioritizing eco-friendly solutions with reducing chemical pesticide usage. The LPN model and DSS provide a robust, efficient, and sustainable solution for precision agriculture.
\section{Conclusion and Future Work}
This study delivered a lightweight model capable of proper pest detection and a decision support system for advising targeted pesticide solutions. The framework effectively balanced high performance in few-shot scenarios with low computational needs, making it practical for real-time use in resource-limited settings. By combining environmental and crop-specific elements, it encouraged sustainable pest management practices. Future measures will aim to enhance the dataset with more pest species, incorporate real-time environmental data via IoT devices, and extend the framework to support a broader variety of crops and agricultural conditions.
\section*{Acknowledgement}
The first author sincerely acknowledges the financial support received from the DST-INSPIRE Fellowship, which has helped him conduct this research.
\bibliographystyle{splncs04} 
\bibliography{reff}   
\end{document}